\documentclass{IOS-Book-Article}
\usepackage[inline]{enumitem}
\usepackage{graphicx}
\usepackage{amsmath}
\usepackage{mathptmx}
\usepackage{soul}\setuldepth{article}
\usepackage{caption}
\usepackage{subcaption}
\usepackage{amssymb}
\usepackage{amsmath}
\def\hb{\hbox to 11.5 cm{}}

\begin{document}

\pagestyle{headings}
\def\thepage{}
\begin{frontmatter}              

\title{MVSBoost: An Efficient Point Cloud-based 3D Reconstruction}

\markboth{}{May 2024\hb}

\author[A]{\snm{Umair Haroon}\orcid{0000-0002-1449-1838}%
\thanks{Corresponding Author: Umair Haroon, umairharoon3797@gmail.com}},
\author[A]{\snm{Ahmad AlMughrabi}\orcid{0000-0002-9336-3200}}
\author[A,B]{\snm{Ricardo Marques}\orcid{0000-0001-8261-4409}}
and
\author[A,C]{\snm{Petia Radeva}\orcid{0000-0003-0047-5172}}

\runningauthor{B.P. Manager et al.}
\address[A]{Universitat de Barcelona}
\address[B]{Computer Vision Center}
\address[C]{Institut de Neurosciències}

\begin{abstract}
Efficient and accurate 3D reconstruction is crucial for various applications, including augmented and virtual reality, medical imaging, and cinematic special effects. While traditional Multi-View Stereo (MVS) systems have been fundamental in these applications, using neural implicit fields in implicit 3D scene modeling has introduced new possibilities for handling complex topologies and continuous surfaces. However, neural implicit fields often suffer from computational inefficiencies, overfitting, and heavy reliance on data quality, limiting their practical use. This paper presents an enhanced MVS framework that integrates multi-view 360-degree imagery with robust camera pose estimation via Structure from Motion (SfM) and advanced image processing for point cloud densification, mesh reconstruction, and texturing. Our approach significantly improves upon traditional MVS methods, offering superior accuracy and precision as validated using Chamfer distance metrics on the Realistic Synthetic 360 dataset. The developed MVS technique enhances the detail and clarity of 3D reconstructions and demonstrates superior computational efficiency and robustness in complex scene reconstruction, effectively handling occlusions and varying viewpoints. These improvements suggest that our MVS framework can compete with and potentially exceed current state-of-the-art neural implicit field methods, especially in scenarios requiring real-time processing and scalability.
\end{abstract}

\begin{keyword}
3D Reconstruction \sep Muti-View Stereo \sep Neural Implicit Fields

\end{keyword}
\end{frontmatter}
\markboth{May 2024\hb}{May 2024\hb}

\section{Introduction}
The pursuit of realistic and accurate 3D reconstruction has long been a focal point in the field of computer vision and graphics, finding applications across a spectrum of industries, including augmented and virtual reality (AR/VR), medical imaging, and special effects in media production. Historically, the development and implementation of Multi-View Stereo (MVS) systems \cite{huang2018deepmvs, zhang2020visibility} have been central to these efforts. MVS, leveraging photometric consistency \cite{kim2022infonerf}, has enabled the reconstruction of complex scenes from multiple images by assessing and integrating the geometric and photometric information captured from different viewpoints.

Despite the robustness and extensive application of MVS in various scenarios, the advent of neural implicit fields \cite{darmon2022improving, xu2022point} has ushered in a transformative era in 3D scene reconstruction. These methods utilize deep learning to create implicit models of scenes, essentially learning continuous volumetric fields that represent complex surfaces and structures. While they offer the advantage of handling intricate topologies and continuous surface representations without the need for explicit mesh construction, they are not without limitations. Neural implicit fields often require extensive computational resources for training and are sensitive to the diversity and quality of training data, issues that can lead to overfitting and generalized inaccuracies in practical applications.

Given the computational inefficiencies and practical limitations of neural implicit fields, there remains a significant opportunity to advance MVS techniques, especially in contexts demanding high precision, real-time processing, and efficient resource utilization. This paper introduces an enhanced MVS framework that significantly refines the traditional approach by integrating the latest advancements in camera pose estimation and sophisticated image processing techniques. By employing a structured pipeline that includes camera pose estimation via Structure from Motion (SfM), followed by point cloud densification, mesh reconstruction, and texture mapping, our method retains and enhances the practical virtues of MVS.

Our comparative studies, grounded in rigorous evaluations using the Realistic Synthetic 360 dataset \cite{mildenhall2021nerf} and Chamfer distance metrics, demonstrate the superior performance of our proposed method over existing state-of-the-art techniques, including those based on neural implicit fields. This paper details a novel MVS-based approach that effectively combines scalability, accuracy, and computational efficiency.  

The remainder of this paper is organized as follows: Section 2 reviews the relevant literature, Section 3 describes our methodology, Section 4 discusses the experimental setup and presents our results, and Section 5 concludes with a summary of our findings and discusses avenues for future research.

%
\section{Related Work}
Pursuing realistic and accurate 3D reconstruction has been a focal point in computer vision and graphics, with applications in augmented and virtual reality, medical imaging, and media production. MVS systems and neural implicit fields have significantly contributed to advancing 3D reconstruction techniques. Moreover, despite these advancements, time complexity, detailed and accurate reconstruction, and the training time for neural implicit fields are still in open discussion.

MVS is a well-established technique to reconstruct dense 3D scene representations from overlapping images. Traditional MVS methods typically rely on comparing RGB image patches using metrics such as Normalized Cross-Correlation (NCC), Sum of Squared Differences (SSD), or Sum of Absolute Differences (SAD) \cite{Scharstein2002}. In recent years, PatchMatch-based MVS approaches \cite{Barnes2009} have become popular due to their high parallelism and robust performance \cite{Galliani2015, Zheng2014}.

The emergence of deep learning has resulted in significant progress in MVS. MVSNet \cite{yao2018mvsnet} constructs cost volumes by warping feature maps from neighboring views and utilizes 3D Convolutional Neural Networks (CNNs) to regularize the cost volumes. To address the memory consumption of 3D CNNs, R-MVSNet \cite{yao2019recurrent} sequentially regularizes 2D cost maps using a gated recurrent network. Other approaches \cite{chen2019point,gu2020cascade} incorporate coarse-to-fine multi-stage strategies to refine the 3D cost volumes progressively. PatchmatchNet \cite{wang2021patchmatchnet} introduces an iterative multiscale PatchMatch strategy in a differentiable MVS architecture. More recently, TransMVSNet \cite{ding2021transmvsnet} integrates Transformers to aggregate long-range context information within and across images.
Still, matching pixels in low-texture or non-Lambertian areas remains challenging, and errors can accumulate from the subsequent point cloud fusion and surface reconstruction steps. 

\subsection{Neural Implicit Fields}
In recent years, neural implicit fields have become increasingly popular for multi-view 3D surface reconstruction. The reason for this growth can be attributed to the introduction of differentiable rendering of implicit functions training methods. In early works, surface rendering procedures were relied upon, wherein the color of a pixel was estimated using the radiance of a single point in the volume \cite{niemeyer2020differentiable, Sitzmann2019}. However, these methods have been surpassed by training procedures based on volume rendering, with multiple samples being taken via ray marching.

The technique of volumetric ray marching, introduced in the pioneering research on Neural Radiance Fields (NeRFs) \cite{Mildenhall2020}, has been adapted to surface modelling, leading to significant improvements in reconstruction quality. This method estimates the color along the ray using the volume rendering integral, approximated as a sum of weighted radiances at multiple points throughout the volume. To increase the accuracy of the approximation, researchers have employed methods based on importance \cite{Wang2021, almughrabi2023pre}, uncertainty \cite{Yariv2021}, or surface intersection-based \cite{Oechsle2021} sampling, which have outperformed simpler strategies such as uniform sampling.

Recent advancements in neural implicit fields have introduced hybrid surface representations that enhance ray marching \cite{dogaru2023sphere}. These representations limit the sampling space to a volume that roughly encompasses the scene in conjunction with the method of the base neural reconstruction. Researchers have optimized the selection of samples surrounding the reconstructed surface \cite{dogaru2023sphere}, which led to a superior reconstruction. Moreover, these hybrid surface representations have been used to guide the sampling of training rays, resulting in better reconstructions within the same training time. The progress in neural implicit fields has been driven by the development of more accurate and efficient sampling strategies, along with integrating these hybrid surface representations. 

The main contributions of our paper are as follows:
\begin{enumerate*}

\item \textbf{We propose a robust MVS framework}  that integrates multi-view $360^0$ imagery with advanced processing techniques to produce detailed and accurate 3D point clouds and meshes;
\item \textbf{We achieve superior performance in terms of accuracy and precision},  outperforming existing neural implicit field approaches and traditional MVS methods, as demonstrated through extensive testing with Chamfer distance metrics on a comprehensive dataset.
\item \textbf{We demonstrate a high computational efficiency} by optimizing each stage of the MVS process, leading to a substantial reduction of the computational resources required, facilitating faster processing and greater efficiency;
\item \textbf{We illustrate the robustness of our approach on complex scene reconstruction}: since our method exhibits exceptional capability in handling occlusions and adapting to varying viewpoints, ensuring reliable and comprehensive scene reconstructions.
\item \textbf{Our method achieves scalability and application versatility:} designed to accommodate both bounded and unbounded scenes without specific image-capturing patterns, the proposed framework adapts seamlessly across various scales and environments, making it suitable for various applications.

\end{enumerate*}


\section{Proposed Methodology}
\label{sec:methodology}
Our study focuses on achieving precise point cloud-based 3D reconstruction of objects from multiple views. This section introduces our novel pipeline, which embodies our key contributions to the field. Central to our approach is the innovative integration of camera pose estimation (SfM) with subsequent stages: Densify Point Cloud, Mesh Reconstruction, Mesh Refinement, and Texture Mesh. By meticulously coordinating these modules, we have achieved superior performance in terms of accuracy and precision compared to existing state-of-the-art methods. Our pipeline not only enhances computational efficiency but also demonstrates robust capability in handling complex scene reconstruction.

\vspace*{5mm}
\begin{figure*}[htb]
\begin{center}
\includegraphics[width=1\linewidth, height=7cm]{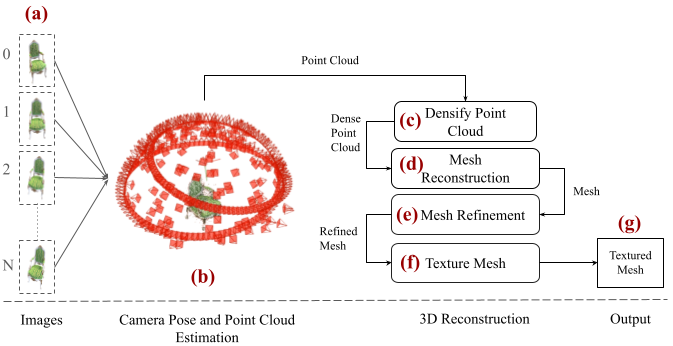}
\end{center}
\vspace*{-5mm}
   \caption{This figure illustrates the proposed multi-phases framework, where (a) takes multi-view 360-degree images with transparent background (i.e; object-centric RGBA images); pass them to (b) Camera Pose and Point Cloud Estimation phase, which estimates a point cloud and camera poses based on the extracted SIFT features from RGBA inputs; the point cloud passed to (c) Densify Point Cloud for obtaining a complete and accurate as possible point-cloud; (d) Mesh Reconstruction for estimating a mesh surface that explains the best the input point-cloud; (e) Mesh Refinement for recovering all fine details; (f) Texture Mesh for computing a sharp and accurate texture to color the mesh; finally (j) is the output mesh.}
\label{fig:methodology}
\end{figure*}

\vspace*{-5mm}
\subsection{Overview}
Our framework aims to achieve high-quality 3D reconstruction of objects in real scenarios by a given set of RGB images.  Our framework  relies on three distinct phases: 1. Input Images, 2. Camera and Point Cloud Estimation, and 3. 3D Reconstruction.

Fig. \ref{fig:methodology} shows a diagram representing the general overview of our proposed framework. In the first phase, (Fig. \ref{fig:methodology}(a)), our framework accepts a $360^0$ scene as a set of input RGBA images. The next phase (Fig. \ref{fig:methodology}(b)) extracts the SIFT features from the given images, triangulates matches of similar features where it helps to estimate the camera locations (i.e. poses), and returns a point cloud. The last phase (Fig. \ref{fig:methodology}(c-f)) relies on four distinct phases. In Densify Point Cloud (Fig. \ref{fig:methodology}(c)), our  framework obtains a complete and accurate possible point cloud from the given point cloud by recovering the missing parts of the scene using a Patch-Match \cite{Barnes2009} approach. In the Mesh Reconstruction (Fig. \ref{fig:methodology}(d)) phase, our  framework uses the dense point cloud obtained from the previous steps as its input resulting in a rough mesh. In the Mesh Refinement phase (Fig. \ref{fig:methodology}(e)), the rough mesh is refined to recover fine details and larger missing parts of objects. Finally, in the Texture Mesh phase (Fig. \ref{fig:methodology}(f)), the mesh is colored based on the input images.


\subsection{Our Proposal: MVSBoost}
\label{sec:nonerf}
Our framework aims to propose rich and expressive meshes from a set of $N_{I}$ input images $\mathcal{I} = \{I _{i}| i = 1 \ldots N_{I}\}$. In this section, we present the Camera Pose and Point Cloud Estimation. Secondly, we present the Densify Point Cloud, Mesh Reconstruction, Mesh Refinement, and Texture Mesh as 3D reconstruction using Point Cloud phases.

\subsection{Camera Pose and Point Cloud Estimation}
SfM \cite{schonberger2016structure} is a sequential process that reconstructs 3D structures from multiple images taken from different viewpoints. It typically involves feature extraction, feature matching, geometric verification, and an iterative reconstruction stage. The initial step involves searching for correspondences to identify scene overlap in a set of input images $\mathcal{I}$ and determining the projections of common points in overlapping images. This process results in a collection of geometrically verified image pairs $\bar{C}$ and a graph representing the projections of each point in the images. 
In the \textbf{feature extraction} step, SfM involves detecting local features $\mathcal{F}_{i} = \{(x_{j} ,f_{j}) | j = 1...N_{F_{i}}\}$ in each image $I_{i}$, where $x_{j}$ represents the location and $f_{j}$ is the appearance descriptor. These features should be invariant to radiometric and geometric changes to ensure unique recognition across multiple images. SIFT and its derivatives and learned features are considered the most robust options, while binary features offer better efficiency at the cost of reduced robustness. 
In the \textbf{feature matching} step, SfM identifies images capturing the same scene elements by utilizing features $\mathcal{F}_{i}$ as descriptors of image appearances. Instead of exhaustively comparing all image pairs for scene overlap, SfM matches features between images $I_{a}$ and $I_{b}$ by assessing the similarity of feature appearances, $f_{j}$. This direct approach has a computational complexity of $O(N_{I}^2 \times N_{F_{i}}^2)$, which becomes impractical for large image datasets. Various methods have been developed to address the challenge of efficient and scalable matching. The outcome is a set of potentially overlapping image pairs $C =\{ \{I_{a}, I_{b}\} | I_{a}, I_{b} \in \mathcal{I}, a < b \}$ along with their corresponding feature matches $M_{ab} \in \mathcal{F}_{a} \times \mathcal{F}_{b}$. 
In the \textbf{geometric verification} step, the third stage in SfM verifies potentially overlapping image pairs ${C}$ by estimating transformations using projective geometry to ensure corresponding features map to the same scene point. Different mappings, such as homography \textbf{H} for planar scenes, epipolar geometry with essential matrix \textbf{E} or fundamental matrix \textbf{F} for moving cameras, and trifocal tensor for three views, are used based on the spatial configuration. Robust techniques like RANSAC handle outlier-contaminated matches, resulting in geometrically verified image pairs $\bar{C}$ with inlier correspondences $\bar{M}_{ab}$ and their geometric relations $G_{ab}$. Decision criteria like GRIC or methods like QDEGSAC aid in determining the appropriate relation, forming a scene graph with images as nodes and verified pairs as edges, resulting in a point cloud data structure, as shown in Fig. \ref{fig:methodology}(b).

\subsection{3D Reconstruction}
\begin{figure}[htb]
\vspace*{-1cm}
    \centering\setlength{\tabcolsep}{1pt} 
    \begin{center}
\begin{tabular}{cccc}
    \begin{subfigure}[b]{.26\linewidth}
      \captionsetup[figure]{font=footnotesize,labelfont=footnotesize}
         \centering
         \includegraphics[width=\textwidth]{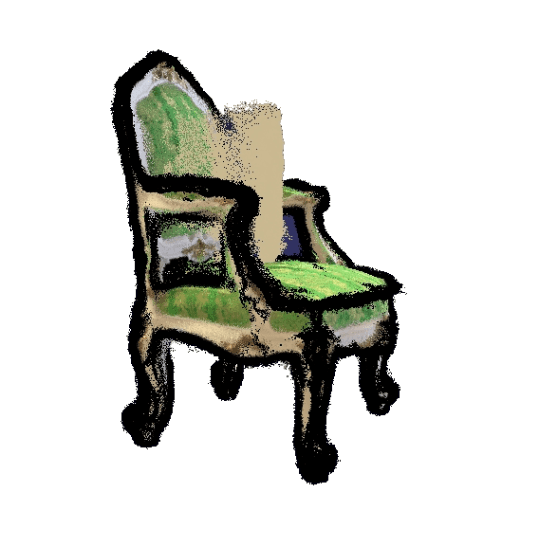}
         \caption{Densify Point Cloud}
         \label{fig:scene55_3v_in}
     \end{subfigure}
      &
      \begin{subfigure}[b]{.26\linewidth}
      \captionsetup[figure]{font=footnotesize,labelfont=footnotesize}
         \centering
         \includegraphics[width=\textwidth]{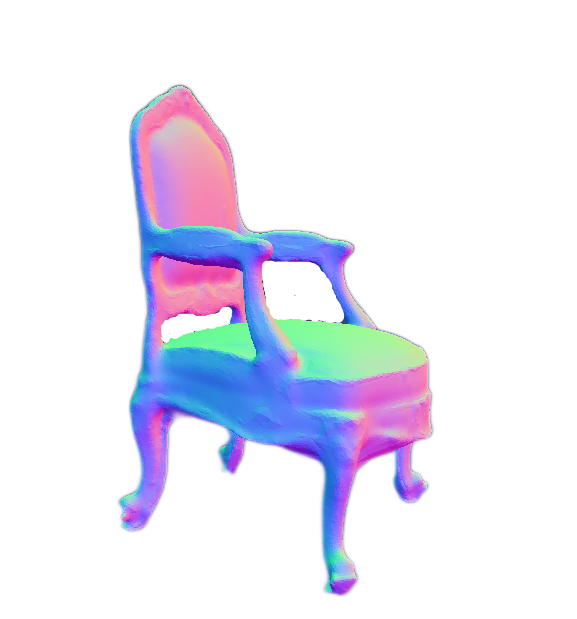}
         \setcounter{subfigure}{1}%
         \caption{Mesh Reconstruction}
         \label{fig:scene55_3v_a}
     \end{subfigure} 
     &
     \begin{subfigure}[b]{.26\linewidth}
         \centering
         \includegraphics[width=\textwidth]{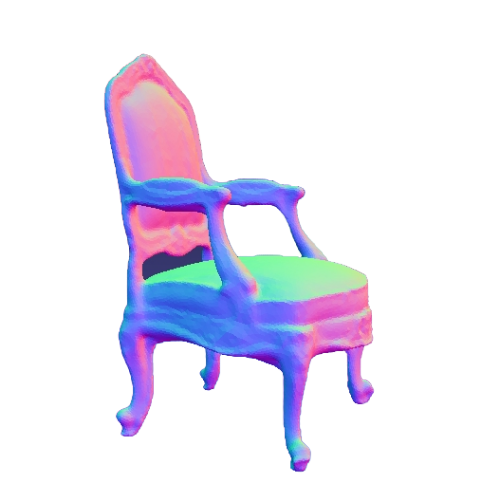}
         \setcounter{subfigure}{2}%
         \caption{Mesh Refinement}
         \label{fig:scene55_3v_b}
     \end{subfigure} 
     &
     \begin{subfigure}[b]{.26\linewidth}
         \centering
         \includegraphics[width=\textwidth]{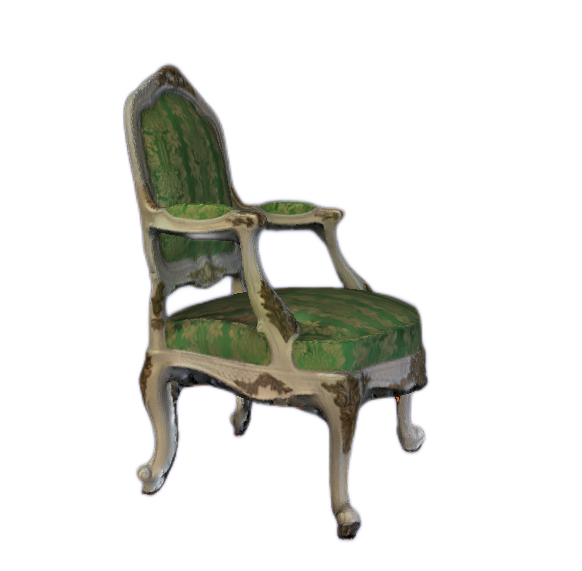}
         \setcounter{subfigure}{3}%
         \caption{Texture Mesh}
         \label{fig:scene55_3v_c}
     \end{subfigure}
    
\end{tabular}
\vspace*{-5mm}
        \caption{An illustration on the 3D reconstruction module on the Chair scene.}
        \label{fig:3d_reconstruction}
        \end{center}
\end{figure}

In \textbf{Densify Point Cloud}, the dense point cloud extraction involves the process of enhancing a sparse point cloud created through SfM by utilizing depth map computation and Depth map fusion techniques. Depth map fusion works by generating depth maps for individual input images through feature matching across adjacent perspectives; the feature matching is conducted at various resolution levels until a detailed set of dense depth maps is obtained. Let $\mathrm{D}_{i}$ represent the depth map for image $I_{i}$, computed through feature matching: $\mathrm{D}_{i} = f(I_{i}, I_{adj}, R) $. Where $I_{adj}$ denotes the set of adjacent images and $R$ represents the resolution and different levels of detail. Subsequently, these depth maps are consolidated into a unified dense point cloud $P_{d}$, effectively eliminating superfluous points and noise during the fusion process: $ P_{d} = g(\mathrm{D}_{i}) $, as shown in Fig. \ref{fig:3d_reconstruction} (a).


\noindent\textbf{Mesh reconstruction} uses an MVS approach to create a mesh from a dense point cloud. The point cloud is first transformed into a tetrahedral mesh $T$ through $f^D$ Delaunay triangulation function: $ T = f^D(P_{d})$. Then, a graph-cut optimization $f^G$ determines whether each tetrahedron is inside or outside the object: $L = f^G(T)$. Finally, the marching cubes algorithm $\mathbb{M}$ extracts the mesh surface from the labeled tetrahedra: $M = \mathbb{M}(T, L)$. The resulting mesh $M$ is a seamless and accurate representation of the object's geometry based on the original point cloud data, as shown in Fig. \ref{fig:3d_reconstruction} (b).


\noindent\textbf{Mesh refinement} is a set of techniques used to improve the quality of reconstructed meshes. These techniques include mesh simplification $f^S$, which reduces the number of vertices while preserving important details: $M_{s} = f^S(M)$, and mesh smoothing $f^S_m$, achieved using Laplacian or bilateral filtering to eliminate noise and outliers: $M_{sm} = f^S_m(M_{s})$. Additionally, mesh denoising $f^D_e$ is used further to enhance the mesh quality through normal voting tensor filtering: $M_{d} = f^D_e(M_{sm})$. Mesh optimization techniques $\hat{f}$, such as vertex relaxation and edge flipping, are employed to refine the quality of the triangles: $\hat{M} = \hat{f}(M_{d})$. These refinement processes create a clean and precise mesh $\hat{M}$ that accurately represents the object's surface characteristics, as shown in Fig. \ref{fig:3d_reconstruction} (c).


\noindent\textbf{Texture mesh} involves mapping images onto the 3D model to create a realistic and detailed surface representation. This process significantly enhances the visual quality of the 3D model, making it suitable for various applications like virtual reality, gaming, and visualization. Texture mapping involves associating 2D image coordinates (texture coordinates) with the vertices of the 3D mesh: $T_{c} = \{(u_{i},v_{i}) | i = 1,..., N_{v}\}$. Apply the texture coordinates $f^T$ to the mesh $\hat{M}$ to obtain the textured mesh: $M_{t} = f^T(\hat{M}, T_{c}, \mathcal{I})$. Texture mapping is crucial for adding color, patterns, and details to the mesh, providing a more immersive and realistic experience when interacting with the 3D model, as shown in Fig. \ref{fig:3d_reconstruction} (d).


\section{Experimental Results}
\label{sec:results}
 To evaluate our proposed approach, we use the Chamfer distance \cite{dogaru2023sphere} metric, a commonly used measure for assessing the quality of surface reconstruction. This metric calculates the average distance between each point in one point set (e.g., the reconstructed surface) and its nearest neighbor in another point set (e.g., the ground truth surface). By combining this distance metric with the Realistic Synthetic 360 dataset, we can comprehensively assess the performance of our proposed method for 3D reconstruction tasks. These benchmark datasets provide a standardized and challenging environment to test the accuracy and robustness of our approach, ensuring that our findings are reliable and comparable to other state-of-the-art methods in the field, as shown in Table. \ref{table:quantitative_comparison_RS360}.

\subsection{Evaluation protocol}
We followed the same evaluation protocol of Sphere NeuralWarp \cite{dogaru2023sphere} for a fair comparison. Notably, since the Chamfer distance is sensitive to rotation, translation, and orientation between meshes, the baselines use extra information to register the predicted mesh with the ground truth mesh such as camera intrinsics information. In contrast, we use Iterative Closest Point (ICP) \cite{rusinkiewicz2001efficient} to rigidly align the groundtruth and the reconstructed mesh avoiding the need of additional information such as camera intrinsics. 

\subsection{Implementation settings}
We used GeForce GTX 1080 Ti/12G to run our experiments. For the Camera Pose and Point Cloud estimation phase (Fig. \ref{fig:methodology}(a)), we use COLMAP \cite{schoenberger2016sfm, schoenberger2016mvs}. For the feature extractor, we set a single camera option, maximum image size as $1000$, default focal length as $5.2$, maximum number of SIFT features as $2048$, BA global function tolerance as $0.000001$, and focal length ratio in range $[0.1\dots10]$.
For the 3D reconstruction phase (Fig. \ref{fig:methodology}(c-f)), we use OpenMVS \cite{cernea2020openmvs}. We set the close-holes parameter as $400$, mesh smooth parameter as $5$, and the maximum resolution parameter as $512$.

\subsection{Datasets}
Our primary experiments use the widely recognized 3D reconstruction benchmark datasets, Realistic Synthetic 360 \cite{Mildenhall2020}. \textbf{The Realistic Synthetic 360 dataset} was created for novel view synthesis \cite{Mildenhall2020}. Recently, it has been repurposed to evaluate 3D reconstruction algorithms. The dataset contains eight scenes (400 images per scene; each image is $800 \times 800$ px, including masks and depth images) with complex geometries and non-Lambertian materials, making it a challenging benchmark. The ground truth meshes are filtered to remove non-visible internal surfaces to ensure fair comparisons. Similarly, the reconstructed meshes also undergo a similar process. 



\subsection{Comparative Analysis}
We compared our method to seven state-of-the-art methods on eight scenes of the Realistic Synthetic 360 dataset, each presenting unique structural complexities in the reconstructed surface. These complexities challenge standard methods in accurately estimating the locations of high-density regions. This limitation hinders the optimization process and diminishes reconstruction quality and rendering performance. 

\noindent\textbf{Quantitative Results:} Our proposed method was evaluated against seven state-of-the-art methods (COLMAP \cite{schonberger2016pixelwise}, NeuS \cite{wang2021neus}, Sphere NeuS \cite{dogaru2023sphere}, NeuS w/ masks \cite{wang2021neus}, Sphere NeuS w/ masks \cite{dogaru2023sphere}, NeuralWarp \cite{darmon2022improving}, and Sphere NeuralWarp \cite{dogaru2023sphere}) using the Realistic Synthetic 360 dataset. The results are shown in Table \ref{table:quantitative_comparison_RS360}, indicating that our method significantly outperforms all scenes compared to the state-of-the-art.

\begin{table}[htb]
\begin{center}
\begin{tabular}{|l|cccccccc|c|}
\hline
Methods&Chair&Drums&Ficus&Hotdog&Lego&Mats&Mic&Ship&Mean  \\
\hline\hline
COLMAP \cite{schonberger2016pixelwise} & 0.77 & 1.26 & 0.96 & 1.95 & 1.36 & 2.19 & 1.33 & 1.00 & 1.42 \\
NeuS \cite{wang2021neus} & 0.38 & 1.88 & 0.51 & 0.52 & 0.68 & 0.40 & 0.60 & 0.60 & 0.70  \\
Sphere NeuS \cite{dogaru2023sphere} & 0.39 & 1.20 & 0.40 & 0.57 & 0.61 & 0.31 & 0.67 & 0.54 & 0.59 \\
NeuS w/ masks \cite{wang2021neus} & 0.40 & 0.90 & 0.41 & 0.58 & 0.67 & 0.28 & 0.59 & 0.73 & 0.57 \\
Sphere NeuS w/ masks \cite{dogaru2023sphere} & 0.45 & 0.94 & 0.32 & 0.54 & 0.67 & 0.27 & 0.57 & 0.71 & 0.56 \\
NeuralWarp \cite{darmon2022improving} & 0.43 & 3.00 & 0.94 & 1.65 & 0.81 & 1.02 & 0.75 & 1.27 & 1.23 \\
Sphere NeuralWarp \cite{dogaru2023sphere} & 0.41 & 2.67 & 0.61 & 1.44 & 0.76 & 0.92 & 0.80 & 1.07 & 1.09 \\
\textbf{Ours} & \textbf{0.22} & \textbf{0.16} & \textbf{0.17} & \textbf{0.26} & \textbf{0.06} & \textbf{0.07} & \textbf{0.22} & \textbf{0.16} & \textbf{0.16} \\
 \hline
\end{tabular}
\end{center}
\caption{A quantitative comparison of our proposed method with the state-of-the-art methods on the Realistic Synthetic 360 dataset using Chamfer distance (the lower the better).} 
\label{table:quantitative_comparison_RS360}
\end{table}

\begin{figure*}
\begin{center}
\includegraphics[width=1\linewidth]{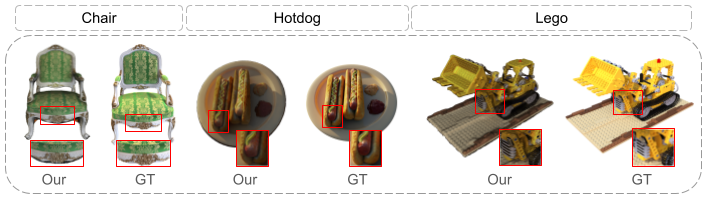}
\end{center}
\vspace*{-5mm}
   \caption{A side-by-side comparison of our method's results with the ground truth from the Realistic Synthetic 360 dataset. The figure illustrates significant differences and improvements in the presented scenes: the Chair, Hotdog, and Ship. Our reconstruction exhibits superior precision, refinement, detail, and texture fidelity compared to existing methods, as evidenced by the clear visual distinctions.}
\label{fig:short}
\end{figure*}

\noindent \textbf{Qualitative Results:} Fig. \ref{fig:short} compares our method with the ground truth on the Realistic Synthetic 360 dataset. It demonstrates that our reconstructions are more accurate and closer to the ground truth than state-of-the-art methods. The dataset provides high-precision 3D surface models, allowing us to evaluate 3D reconstruction algorithms quantitatively. By comparing our reconstructions with the ground truth, we can assess the accuracy and completeness of our results. The figure shows that our method closely matches the ground truth's detailed geometry and surface properties, highlighting its efficiency in reconstructing 3D objects from multi-view images.

\subsection{Discussion}
The results validate our hypothesis that augmenting traditional MVS techniques with modern computational methods and 360-degree camera technologies can surpass recent neural approaches in real-world applications. Our integration of SfM algorithms and 360-degree cameras enhances reconstruction quality, with substantial potential for further optimization in real-time processing and dynamic scene handling.

\noindent \textbf{Complexity:} Our MVS framework optimizes the reconstruction process through GPU acceleration and improved algorithms for image processing and mesh reconstruction, significantly reducing computational time and memory consumption compared to traditional state-of-the-art methods. We re-trained the state-of-the-art approaches on their defined settings using the same machine for a fair comparison. The state-of-the-art methods of the neural implicit field with which we are comparing take considerably longer: Sphere Neus \cite{dogaru2023sphere} takes 4 hours, 17 minutes, and 26.411 seconds, and Sphere NeuralWarp \cite{dogaru2023sphere} takes 9 hours, 29 minutes, and 47.156 seconds. In contrast, our proposed method takes only around 9 minutes and 19.85 seconds. This substantial reduction in computational time proves our claim that our proposed method significantly improves efficiency and minimizes memory consumption, thereby enhancing the system’s capacity to handle larger datasets effectively.

\begin{figure*}
\begin{center}
\includegraphics[width=1\linewidth]{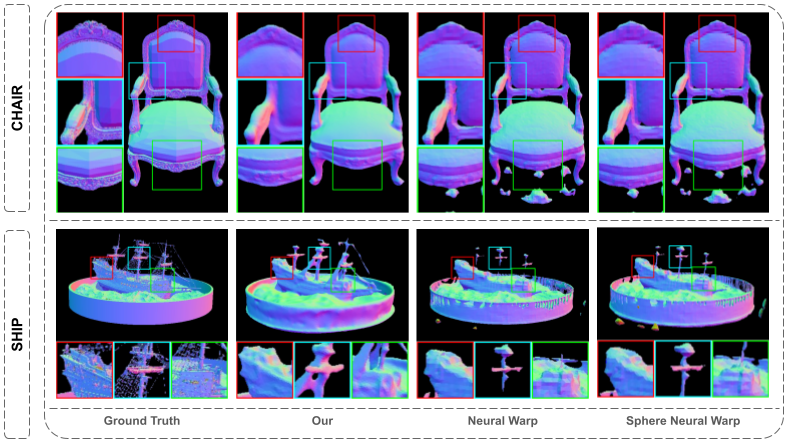}
\end{center}
\vspace*{-5mm}
   \caption{A detailed comparison showcasing selected mesh parts from ground truth, our method, and previous state-of-the-art methods, demonstrating robustness by preserving a high level of details in challenging areas where previous methods struggle, as highlighted in this figure.}

\label{fig:short2}
\end{figure*}

\noindent \textbf{Corner cases:} Our method excels in reconstructing complex scenes with significant occlusions and diverse viewpoints. The robustness of our approach is evident in its consistent detail and accuracy, attributed to comprehensive scene coverage from 360-degree imagery and refined camera pose estimation techniques, as shown in Fig. \ref{fig:short2}.

\noindent \textbf{Limitations:}
Despite its advantages, our method has some limitations:
\begin{enumerate*}
    \item Requires a high number of images for perfect reconstruction. 
    \item Struggles with high-frequency areas such as reflections.
    \item May not perform well under varying lighting conditions.
\end{enumerate*}
%

\section{Conclusions and Future Work}
\label{sec:conclusion}
This study demonstrates the significant advantages of our proposed MVS framework over existing neural implicit field methods. Using the Realistic Synthetic 360 dataset and Chamfer distance metrics, our method proves to be more accurate, precise, and computationally efficient. The integration of multi-view 360-degree imagery with advanced SfM techniques, point cloud densification, mesh reconstruction, and texturing processes results in high-quality 3D models that effectively handle complex scenes, occlusions, and diverse viewpoints. Our findings highlight the practicality of MVS in real-time and large-scale applications, making it suitable for AR/VR, medical imaging, and media production. Future work will aim to refine the MVS process, explore more efficient algorithms for image processing, and extend applicability to more challenging environments, maintaining its relevance and superiority in 3D reconstruction technologies.

\bibliographystyle{IEEEtran}
\bibliography{reference}

\end{document}